\documentclass[letterpaper, 10 pt, conference]{ieeeconf}
\IEEEoverridecommandlockouts
\let\labelindent\relax
\usepackage{enumitem}
\usepackage{balance}
\usepackage[font={small}]{caption}
\usepackage{subcaption}
\usepackage{array}
\usepackage{textcomp}
\usepackage{mathtools, nccmath}
\usepackage{graphicx}
\usepackage{amsfonts}
\usepackage{amsmath}
\usepackage{amssymb}
\usepackage{algorithm}
\usepackage{algorithmic}
\usepackage{hyperref}
\usepackage{tikz}
\usepackage{arydshln}
\usepackage{multirow}
\usepackage{bm}
\usepackage{epstopdf}
\usepackage{cite}
\usepackage{siunitx}

\usepackage{amsthm}

\newtheoremstyle{mystyle}
  {}
  {}
  {}
  {}
  {\bfseries}
  {.}
  { }
  {\thmname{#1}\thmnumber{ #2}\thmnote{ (#3)}}

\theoremstyle{mystyle}

\newtheorem{remark}{Remark}

\newcommand{\optional}[1]{{\color{black} #1}}
\newcommand{\edit}[1]{{\color{black} #1}}

\title{\LARGE \bf Safety-Critical Control and Planning for Obstacle Avoidance between Polytopes with Control Barrier Functions}
\author{Akshay Thirugnanam$^*$, Jun Zeng$^*$, and Koushil Sreenath
\thanks{$^*$Authors have contributed equally and are listed alphabetically.}
\thanks{This work is supported in part by National Science Foundation Grant CMMI-1931853.}
\thanks{All authors are with Hybrid Robotics Group at the Department of Mechanical Engineering, UC Berkeley, USA. \tt\small\{akshay\_t, zengjunsjtu, koushils\}@berkeley.edu}
\thanks{The animation video can be found at \url{https://youtu.be/2hKlihdERog} \edit{and the implementation code can be found at \url{https://github.com/HybridRobotics/cbf}.}}
}
\begin{document}

\maketitle
\begin{abstract}
Obstacle avoidance between polytopes is a challenging topic for optimal control and optimization-based trajectory planning problems. Existing work either solves this problem through mixed-integer optimization, relying on simplification of system dynamics, or through model predictive control with dual variables using distance constraints, requiring long horizons for obstacle avoidance. In either case, the solution can only be applied as an offline planning algorithm. In this paper, we exploit the property that a smaller horizon is sufficient for obstacle avoidance by using discrete-time control barrier function (DCBF) constraints and we propose a novel optimization formulation with dual variables based on DCBFs to generate a collision-free dynamically-feasible trajectory. The proposed optimization formulation has lower computational complexity compared to existing work and can be used as a fast online algorithm for control and planning for general nonlinear dynamical systems. We validate our algorithm on different robot shapes using numerical simulations with a kinematic bicycle model, resulting in successful navigation through maze environments with polytopic obstacles.
\end{abstract}
\section{Introduction}
\label{sec:introduction}
Obstacle avoidance in optimization-based control and trajectory planning has received significant attention in the robotics community.
When a tight-fitting obstacle avoidance motion is expected, the robot and the obstacles need to be considered as polyhedral.
In this paper, we propose an optimization formulation to consider obstacle avoidance between polytopes using discrete-time control barrier function (DCBF) constraints with dual variables.
The proposed formulation is shown to be a computationally fast algorithm that can serve as a local planner to generate dynamically-feasible and collision-free trajectories, or even directly as a safety-critical controller for general dynamical systems.

\begin{figure}[!htp]
	\centering
	\includegraphics[width=1.0\linewidth]{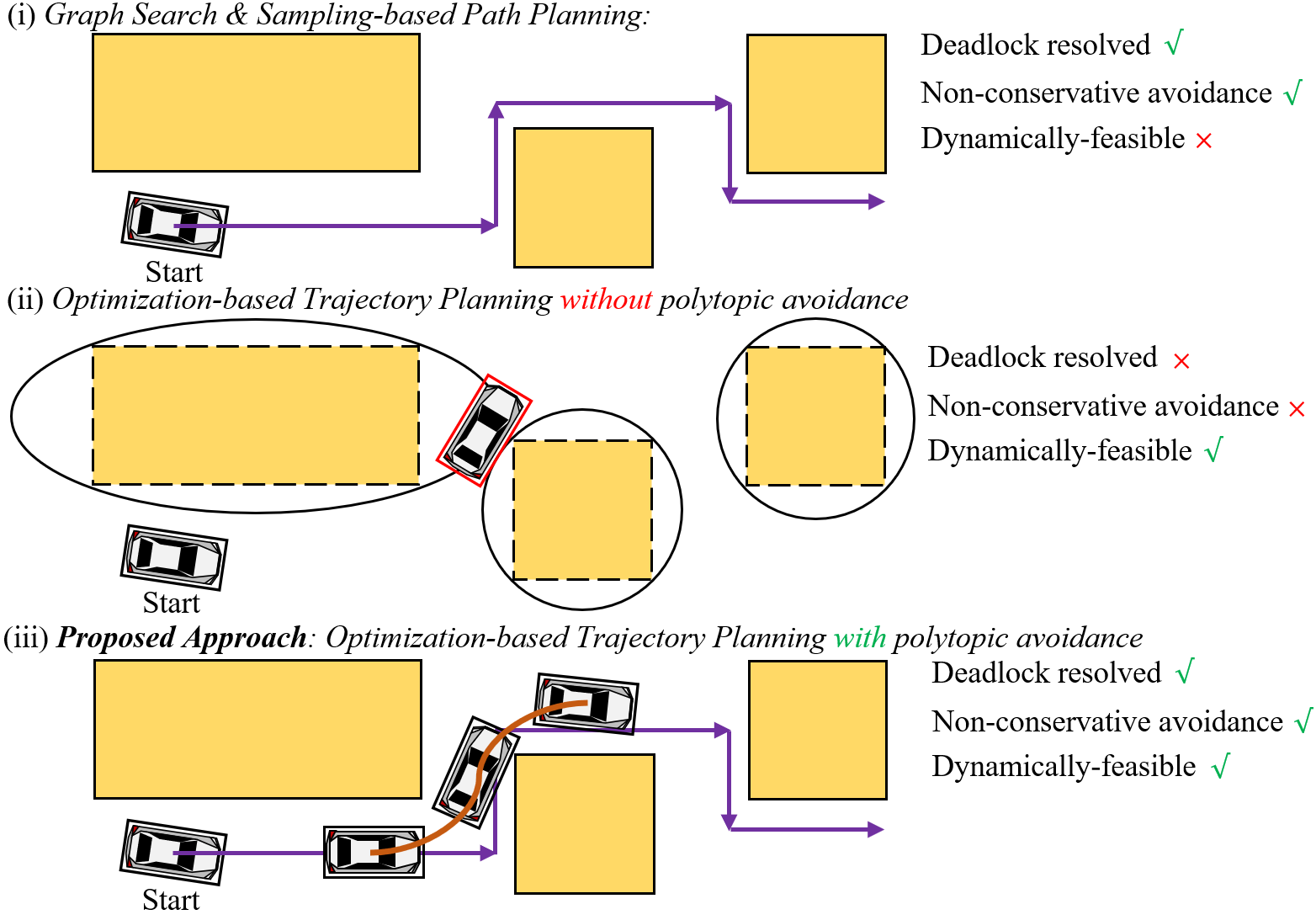}
	\caption{Comparison of approaches using optimization-based trajectory planning with obstacle avoidance.
	The proposed algorithm developed in this paper allows fast optimization, which can be used as an optimal controller or a trajectory planner for general nonlinear systems with obstacle avoidance between polytopes.
	}
    \label{fig:cover}
\end{figure}

\subsection{Related Work}

\subsubsection{Graph Search-based and Sampling-based Approaches}
Motion planning techniques in real-world applications often consider high-level path planning and low-level control synthesis, given safety requirements and dynamical constraints.
Graph search-based and sampling-based approaches such as PRM~\cite{bohlin2000path}, A*~\cite{duchovn2014path}, RRT*~\cite{islam2012rrt} have been explored, and many variant approaches have also been proposed based on them, which could be applied as efficient strategies for high-dimensional kinematic planning.
However, generally, these algorithms assume that a low-level controller exists, and is able to track kinematically feasible trajectories in real time.
This leads to trajectories that are dynamically infeasible and results in large tracking errors on dynamical systems.
Other approaches such as kinodynamic RRT*~\cite{webb2013kinodynamic}, LQR-RRT*~\cite{perez2012lqr} try to bridge the gap between path planning and control synthesis by finding appropriate steering inputs to go between two vertices in the sampling graph.
However, these approaches cannot do dynamic collision checking with respect to the exact nonlinear dynamics of the robot.
For general dynamical systems, we still need to locally generate a dynamically feasible and collision-free trajectory.

\subsubsection{Optimization-based Control and Trajectory Planning}
We now narrow down our discussion to optimization-based approaches for generating collision-free trajectories.
The existing methods in this sub-area can be classified under two categories: those that generate obstacle avoidance behaviors with additional cost terms, and those that apply constraints to achieve a similar behavior.
Additional cost terms were first introduced under the philosophy of potential fields~\cite{hwang1992potential}, and were later generalized to be named as ``barrier function" \cite{wills2004barrier}.
This approach has been applied to solve optimal control and trajectory generation problems with broad applications~\cite{schulman2014motion, liniger2015optimization, romdlony2016stabilization, obeid2018barrier, wu2019control}.
Other methods consider the obstacle avoidance criteria as constraints in the optimization problem.
An example of such a constraint is the distance constraint, enforced using inequality constraints on the distance function between the robot and obstacles, where the robots and the obstacles are usually approximated as points~\cite{patel2011trajectory}, lines~\cite{blackmore2011chance}, paraboloids~\cite{ferraguti2020control}, ellipsoids~\cite{rosolia2016autonomous}, or hyper-spheres~\cite{zeng2021safety}.
The distance functions for these shapes have analytical expressions and are differentiable so that nonlinear optimization (NLP) solvers can easily compute the gradients.

\subsubsection{Obstacle Avoidance between Polytopes}
When a tight-fitting obstacle avoidance motion is expected, the above over-approximations of the shape of the robot can lead to deadlock maneuvers, shown in Fig.~\ref{fig:cover}.
A tight polytopic approximation of the shape of the robot enables obstacle avoidance maneuvers that are less conservative, see \cite{kennedy2018optimal}.
However, the distance function between two polytopes is implicit and not analytic~\cite{grossmann2002review}, and requires a large amount of numerical computation~\cite{gilbert1990computing, deits2015computing}.
Moreover, this distance function between polytopes is also non-differentiable with respect to the robot's configuration, which makes it hard to be treated as a constraint directly for a nonlinear programming problem.
For collision avoidance between two polytopes, mixed-integer programming~\cite{grossmann2002review, richards2002aircraft, deits2015efficient} applies well for linear systems but cannot be deployed as a real-time controller or trajectory planner for general nonlinear systems due to the added complexity from integer variables~\cite{liu2010convex}.
To handle the non-differentiability of the distance function between two polytopes, a duality-based approach~\cite{zhang2018autonomous} is introduced to reformulate constraints as a set of smooth non-convex ones.
However, obstacle avoidance behavior with this method can only be achieved with a relatively long horizon and needs to be solved offline for nonlinear systems~\cite{zhang2020optimization, zeng2020differential, shen2020collision, firoozi2020distributed, gilroy2021autonomous}.
Recently, a dual optimization formulation~\cite{thirugnanam2021duality} was introduced to construct a differentiable control barrier function (CBF)~\cite{ames2019control} for polytopes, but it only optimizes one-step control input and is only applicable for continuous-time affine systems with a relative-degree of one.
This formulation could also run into a deadlock for general high relative-degree systems.

\subsubsection{Obstacle Avoidance with DCBFs}
To resolve the problems mentioned above, it's required to propose a computationally fast multi-step optimization formulation for systems with nonlinear discrete-time dynamics.
Recently, it has been shown that considering discrete-time control barrier function (DCBF) constraints instead of distance constraints can handle this challenge, where the DCBF constraints can regulate the obstacle avoidance behavior with a smaller horizon and prevent local deadlock in trajectory generation, see \cite{zeng2021enhancing}.
The control and planning problems with one-step~\cite{agrawal2017discrete} or multi-step~\cite{zeng2021enhancing} optimization using DCBF constraints have been studied, and various applications on different platforms, including car racing~\cite{zeng2021safety}, autonomous vehicles~\cite{ma2021model}, and bipedal robots~\cite{teng2021toward} have validated this approach.
In the work mentioned above, the robots and the obstacles are only considered as points or hyper-spheres, while obstacle avoidance constraint between polytopes is still an unsolved problem in all previous work by using discrete-time control barrier functions.

\subsection{Contributions}
The contributions of our paper are as follows:
\begin{itemize}
    \item We formulate the dual form of the obstacle avoidance constraint between polytopes as DCBF constraints for safety.
    These proposed DCBF constraints are incorporated into an NMPC formulation which enables  fast online computation for control and planning for general nonlinear dynamical systems.
    \item The proposed NMPC-DCBF formulation for polytopes is validated numerically. 
    Different convex and non-convex shaped robots are shown to be able to navigate with tight maneuvers through maze environments with polytopic obstacles using fast real-time control and trajectory generation.
\end{itemize}


\section{Background}
\label{sec:background}
In this section, we present a brief background on optimization formulations using discrete-time control barrier functions and obstacle avoidance between polytopic sets.
\subsection{Optimization Formulation using DCBFs}
Consider a discrete-time dynamical system with states $x \in \mathcal{X} \subset \mathbb{R}^n$ and inputs $u \in \mathcal{U} \subset \mathbb{R}^m$, as
\begin{equation}
\label{eq:dt-system}
x_{k+1} = f(x_k, u_k),
\end{equation}
where $x_{k} := x(k)$, $u_{k} := u(k)$, $k \in \mathbb{Z}^+$, $\mathcal{U}$ is a compact set and $f$ is continuous.

\subsubsection{Discrete-time CBFs}
Obstacle avoidance for safety for this dynamical system is defined in terms of invariance of its trajectories with respect to a connected set.
In other words, if the dynamical system \eqref{eq:dt-system} is safe with respect to a set $\mathcal{S} \subset \mathcal{X}$, then any trajectory starting inside $\mathcal{S}$ remains inside $\mathcal{S}$.
The set $\mathcal{S}$ is defined as the 0-superlevel set of a continuous function $h: \mathcal{X} \rightarrow \mathbb{R}$ as:
\begin{equation}
\label{eq:safe-set-def}
\mathcal{S} := \{x \in \mathcal{X} \subset \mathbb{R}^n: h(x) \geq 0\}.
\end{equation}
We refer to $\mathcal{S}$ as the safe set and it represents the region outside the obstacle. $h$ is defined as a discrete-time control barrier function (DCBF) if $\forall \; x \in \mathcal{S}, \exists \; u \in \mathcal{U}$ such that
\begin{equation}
\label{eq:dcbf-def}
h(f(x,u)) \geq \gamma(x) h(x), \quad 0 \leq \gamma(x) < 1,
\end{equation}
Let $\gamma_k := \gamma(x_k)$.
Satisfying \eqref{eq:dcbf-def} implies $h(x_{k+1}) \geq \gamma_k h(x_k)$, i.e., the lower bound of the DCBF decreases exponentially with the decay rate $\gamma_k$~\cite{agrawal2017discrete}.
Given a choice of $\gamma(x)$, we denote $\mathcal{K}(x)$ as
\begin{equation}
\label{eq:safe-control-set-def}
\mathcal{K}(x) := \{u \in \mathcal{U}: h(f(x,u)) - \gamma(x) h(x) \geq 0\}.
\end{equation}
Then, if $x_0 \in \mathcal{S}$ and $u_k \in \mathcal{K}(x_k)$, then $x_{k} \in \mathcal{S}$ for $\forall \; k \in \mathbb{Z}^+$, i.e., the resulting trajectory is safe~\cite{agrawal2017discrete}.

Given a valid DCBF $h$~\cite{ames2019control}, imposing DCBF constraint \eqref{eq:dcbf-def} in an optimization problem could guarantee system safety, i.e., collision-free trajectories.
If $\gamma(x)$ is close to $1$, the system converges to $\partial \mathcal{S}$ slowly but can easily become infeasible.
On the other hand, if $\gamma(x)$ is close to $0$, the constraint \eqref{eq:dcbf-def} is feasible in a larger domain but can approach $\partial \mathcal{S}$ quickly and become unsafe.
The later proposed formulation in~\cite{zeng2021enhancing} introduces a relaxing form of DCBF constraint as follows,
\begin{equation}
\label{eq:relaxing-dcbf-def}
h(f(x,u)) \geq \omega(x) \gamma(x) h(x), \quad 0 \leq \gamma(x) < 1.
\end{equation}
where the relaxing variable $\omega$ resolves the tradeoff between feasibility and safety and is optimized with other variables inside an optimization formulation.

When one-step control input is optimized~\cite{agrawal2017discrete}, it could lead to a deadlock situation such that the robot is safe but unable to track the reference command.
A nonlinear model predictive control formulation~\cite{zeng2021safety} can overcome these problems, shown as follows,

\noindent\rule{\columnwidth}{0.4pt}
\textbf{NMPC-DCBF~\cite{zeng2021safety}:}
\begin{subequations}
\label{eq:nmpc-dcbf}
\begin{align}
    & \min_{U, \Omega} p(x_{t+N|t}){+}\sum_{k=0}^{N-1}q(x_{t+k|t},u_{t+k|t}){+}\psi(\omega_k) \label{subeq:nmpc-dcbf-cost}\\
    \text{s.t.} \ & x_{t|t} = x_t, \label{subeq:nmpc-dcbf-initial-condition} \\
    & x_{t+k+1|t} {=} f(x_{t+k|t}, u_{t+k|t}), \quad k{=}0,...,N{-}1\label{subeq:nmpc-dcbf-dynamics-constraint} \\
    & u_{t+k|t} \in \mathcal{U}, \ x_{t+k|t} \in \mathcal{X}, \qquad \; k{=}0,...,N{-}1 \label{subeq:nmpc-dcbf-input-constraint} \\
    & h (x_{t+k+1|t}) \geq  \omega_k \gamma_k h(x_{t+k|t}), \quad \omega_k \geq 0 \notag \\
    & \qquad \text{for}\ k{=}0,...,N_{\text{CBF}}{-}1, \label{subeq:nmpc-dcbf-cbf-constraint}
\end{align}
\end{subequations}
\noindent\rule{\columnwidth}{0.4pt}
\noindent
where $x_{t+i|t}$ and $u_{t+i|t}$ denote the predicted state and input at time $t+i$ evaluated at the current time $t$.
$N$ and $N_{\text{CBF}} \leq N$ denote the prediction and safety horizons respectively, which allows us to control the optimization computation complexity, and $U=[u_{t|t}^T,...,u_{t+N-1|t}^T]^T$ and $\Omega = [\omega_0,...,\omega_{N_{\text{CBF}}-1}]$ are the joint input and relaxation variables respectively.
$p(\cdot)$ and $q(\cdot,\cdot)$ are the terminal and stage costs respectively, and $\psi$ is the penalty function for the relaxation variable.

The optimization formulation \eqref{eq:nmpc-dcbf} can be regarded as a control problem with $U^{*}=[u{^*}_{t|t}^T,...,u{^*}_{t+N-1|t}^T]^T$ as the optimized control inputs, as well as a trajectory planning problem with $X^{*}=[x{^*}_{t|t}^T,...,x{^*}_{t+N|t}^T]^T$ as the optimized trajectory.
From the joint optimal input vector $U^*$ the first input $u^*_{t|t}$ is applied at time $t \in \mathbb{Z}^+$, and the optimization \eqref{eq:nmpc-dcbf} is solved again at time $t+1$ with $x_{t+1}$.

\subsection{Obstacle Avoidance between Polytopic Sets}
In this work, we assume that there are $N_{\mathcal{O}}$ static obstacles together with a single controlled robot.
We further assume that the geometry of all the obstacles and the robot can be over-approximated with a union of convex polytopes, which is defined as a bounded polyhedron.

Let the state of the robot be $x \in \mathbb{R}^n$ with its discrete-time dynamics as defined in \eqref{eq:dt-system} and the geometry of the robot and obstacles be in a $l$-dimensional space.
We denote the geometry of the $i$-th static obstacle and the dynamic robot at some state $x \in \mathcal{X}$ by the polytopes:
\begin{align}
\label{eq:obs-robot-geo-def}
\mathcal{O}_i & := \{ y \in \mathbb{R}^l: A^{\mathcal{O}_i} y \leq b^{\mathcal{O}_i}\} \\ \nonumber
\mathcal{R}(x) & := \{y \in \mathbb{R}^l: A^{\mathcal{R}}(x) y \leq b^{\mathcal{R}} (x) \},
\end{align}
respectively, where $b^{\mathcal{O}_i} \in \mathbb{R}^{s^{\mathcal{O}_i}}, i\in\{1,...N_{\mathcal{O}}\}$ and $b^{\mathcal{R}} (x) \in \mathbb{R}^{s^{\mathcal{R}}}$, and $A^{\mathcal{R}}, b^{\mathcal{R}}$ are continuous.
Inequalities on vectors are enforced element-wise.
We assume that $\mathcal{O}_i$, $i\in\{1,...N_{\mathcal{O}}\}$ and $\mathcal{R}(x) \; \forall \; x \in \mathcal{X}$ are bounded and non-empty.
$s^{\mathcal{O}_i}$, $s^{\mathcal{R}}$ represent the number of facets of polytopic sets for the $i$-th obstacle and the robot, respectively.

Then $\mathcal{O}_i, \; \forall \; i\in\{1,...,N_{\mathcal{O}}\}$, and $\mathcal{R}(x), \; \forall \; x \in \mathcal{X}$, are non-empty, convex, and compact sets, and the minimum distance between any pair $(\mathcal{O}_i,\mathcal{R}(x))$ is well-defined.
The minimum distance is $0$ if and only if $\mathcal{O}_i$ and $\mathcal{R}(x)$ intersect.
The square of the minimum distance between $\mathcal{O}_i$ and $\mathcal{R}(x)$, denoted by $h_i(x)$, can be computed using a QP as follows:
\begin{subequations}
\label{eq:min-dist-primal-qp}
\begin{align}
    h_{i}(x) = & \min_{(y^{\mathcal{O}_i}, y^{\mathcal{R}}) \in \mathbb{R}^{2l}} \quad \lVert y^{\mathcal{O}_i} - y^{\mathcal{R}} \rVert_2^2 \label{subeq:min-dist-primal-qp-cost} \\
    \text{s.t.} \quad & A^{\mathcal{O}_i} y^{\mathcal{O}_i} \leq b^{\mathcal{O}_i},
    \ A^{\mathcal{R}}(x) y^{\mathcal{R}} \leq b^{\mathcal{R}} (x) \label{subeq:min-dist-primal-cons}.
\end{align}
\end{subequations}
where \eqref{eq:min-dist-primal-qp} is a convex optimization problem.
To ensure safe motion of the robot, we enforce DCBF constraints \eqref{eq:dcbf-def} pairwise between each robot-obstacle pair.
Then, the safe set corresponding to the pair $(\mathcal{O}_i,\mathcal{R})$ is defined as:
\begin{equation}
\label{eq:safe-set-pair}
\mathcal{S}_i := \{x \in \mathbb{R}^n: h_i(x) > 0\}^c,
\end{equation}
where $(\cdot)^c$ denotes the closure of a set.
Enforcing the DCBF constraint for each $h_i$ ensures that the state remains in $\mathcal{S}_i$ for all $i$, and thus the state remains in $\mathcal{S} := \cap_{i=1}^{N_\mathcal{O}} S_i$.
Note that, due to the relaxation variable $\omega$, enforcing multiple DCBF constraints for each $h_i$ is equivalent to enforcing a single DCBF constraint on $h(x) := \min_{i} \{h_i(x)\}$.
Thus, we focus on how to enforce the DCBF constraint for a given pair $(\mathcal{O}_i,\mathcal{R})$.

However, \eqref{eq:dcbf-def} requires computation of $h_i(f(x,u))$ via \eqref{eq:min-dist-primal-qp}, which can only be solved numerically.
This results in an optimization formulation with non-differentiable implicit constraints, which results in a significant increase in computation time.
In the following section we derive explicit differentiable constraints which guarantee that the DCBF constraint is satisfied without affecting the feasible set of safe inputs $\mathcal{K}(x) \; \forall \; x \in \mathcal{X}$.

\section{Optimization with Dual DCBF Constraints}
\label{sec:algorithm}
In this section, we derive the duality-based optimization which allows generating obstacle avoidance maneuvers for the controlled robot in tight environments with obstacles.

\subsection{Dual Optimization Problem}
Corresponding to the minimization problem \eqref{eq:min-dist-primal-qp}, we can define a dual problem.
The dual problem is always a convex optimization problem, and a maximization problem if the original primal problem is a minimization one.
The dual formulation of \eqref{eq:min-dist-primal-qp} can be explicitly computed as~\cite[Chap.~8]{boyd2004convex}:
\begin{subequations}
\label{eq:min-dist-dual-prob}
\begin{align}
    g_i(x) = & \max_{(\lambda^{\mathcal{O}_i},\lambda^{\mathcal{R}})} \quad -\lambda^{\mathcal{O}_i} b^{\mathcal{O}_i} - \lambda^{\mathcal{R}} b^{\mathcal{R}} (x) \label{subeq:min-dist-dual-prob-cost} \\
    \text{s.t.} \quad & \lambda^{\mathcal{O}_i} A^{\mathcal{O}_i} + \lambda^{\mathcal{R}} A^{\mathcal{R}}(x) = 0, \label{subeq:min-dist-dual-prob-plane-cons} \\
    & \lVert \lambda^{\mathcal{O}_i} A^{\mathcal{O}_i} \rVert_2 \leq 1, \ \lambda^{\mathcal{O}_i} \geq 0, \ \lambda^{\mathcal{R}} \geq 0 \label{subeq:min-dist-dual-prob-non-neg-cons}.
\end{align}
\end{subequations}
Here $\lambda^{\mathcal{O}_i} A^{\mathcal{O}_i}$ represents the normal vector of the plane of maximum separation between the polytopes.

The Weak Duality Theorem~\cite[Chap.~5]{boyd2004convex} states that $g_i(x) \leq h_i(x)$ holds for all optimization problems.
Since \eqref{eq:min-dist-primal-qp} is a convex optimization with linear constraints and has a well-defined optimum solution in $\mathbb{R}^+$, the Strong Duality Theorem~\cite[Chap.~5]{boyd2004convex} also holds, which states that
\begin{equation}
\label{eq:sdt}
g_i(x) = h_i(x).
\end{equation}

\subsection{Obstacle Avoidance with Dual Variables}
\label{subsec:obs-avoidance-dual-variables}
In order to remove the implicit dependence of $h_i(x)$ on $x$ via \eqref{eq:min-dist-primal-qp}, we enforce a constraint stronger than \eqref{eq:dcbf-def} which does not require explicit computation of $h_i(x)$ via \eqref{eq:min-dist-primal-qp}.
The dual formulation of \eqref{eq:min-dist-primal-qp}, \eqref{eq:min-dist-dual-prob}, can be used to achieve this.

Let $\bar{g}_i(x,\lambda^{\mathcal{O}_i},\lambda^{\mathcal{R}})$ be the cost corresponding to any feasible solution $(\lambda^{\mathcal{O}_i},\lambda^{\mathcal{R}})$ of \eqref{eq:min-dist-dual-prob}.
Since \eqref{eq:min-dist-dual-prob} is a maximization problem and the Strong Duality Theorem \eqref{eq:sdt} holds for the primal problem \eqref{eq:min-dist-primal-qp},
\begin{equation}
\label{eq:dual-feas-ineq}
\bar{g}_i(x,\lambda^{\mathcal{O}_i},\lambda^{\mathcal{R}}) := {-}\lambda^{\mathcal{O}_i} b^{\mathcal{O}_i} {-} \lambda^{\mathcal{R}} b^{\mathcal{R}}(x) \leq h_i(x).
\end{equation}
Then, at time $k$, we can enforce the stronger constraint
\begin{equation}
\label{eq:strong-dcbf-def}
-\lambda^{\mathcal{O}_i} b^{\mathcal{O}_i} - \lambda^{\mathcal{R}} b^{\mathcal{R}}(f(x_k,u)) \geq \gamma_kh_i(x_k)
\end{equation}
which, using \eqref{eq:dual-feas-ineq}, implies
\begin{equation}
\label{eq:strong-dcbf-proof}
    h_i(f(x_k,u)) \geq \gamma_kh_i(x_k),
\end{equation}
as required.
Hence, the DCBF constraint can be enforced by \eqref{eq:strong-dcbf-def} subject to $(f(x_k,u),\lambda^{\mathcal{O}_i},\lambda^{\mathcal{R}})$ being feasible, i.e., the stronger DCBF constraint \eqref{eq:strong-dcbf-def} along with the feasibility constraints \eqref{subeq:min-dist-dual-prob-plane-cons}, \eqref{subeq:min-dist-dual-prob-non-neg-cons} should be satisfied
\begin{subequations}
\label{eq:strong-dcbf-set-constraints}
\begin{align}
    & {-}\lambda^{\mathcal{O}_i} b^{\mathcal{O}_i} {-} \lambda^{\mathcal{R}} b^{\mathcal{R}}(f(x_k,u)) \geq \gamma_kh_i(x_k), \label{subeq:mod-dcbf-controller-cbf-cons} \\
    & \lambda^{\mathcal{O}_i} A^{\mathcal{O}_i} {+} \lambda^{\mathcal{R}} A^{\mathcal{R}}(f(x_k,u)) {=} 0, \label{subeq:mod-dcbf-controller-plane-cons} \\
    & \lVert \lambda^{\mathcal{O}_i} A^{\mathcal{O}_i} \rVert_2 \leq 1, \ \lambda^{\mathcal{O}_i} \geq 0, \ \lambda^{\mathcal{R}} \geq 0 \label{subeq:mod-dcbf-controller-non-neg-cons}.
\end{align}
\end{subequations}

By the Strong Duality Theorem \eqref{eq:sdt}, $\exists \; \lambda^{\mathcal{O}_i*}, \lambda^{\mathcal{R}*}$ satisfying \eqref{subeq:min-dist-dual-prob-plane-cons}-\eqref{subeq:min-dist-dual-prob-non-neg-cons} such that for all $x \in \mathcal{X}$,
\begin{equation}
\label{eq:dual-opt-eq}
\bar{g}_i(x,\lambda^{\mathcal{O}_i*},\lambda^{\mathcal{R}*}) = -\lambda^{\mathcal{O}_i*} b^{\mathcal{O}_i} - \lambda^{\mathcal{R}*} b^{\mathcal{R}}(x) = h_i(x).
\end{equation}
This means that for any fixed $x \in \mathcal{X}$, the input $u$ satisfies the DCBF constraint \eqref{eq:dcbf-def} with the implicit definition of $h_i$ if and only if the tuple $(u,\lambda^{\mathcal{O}_i*},\lambda^{\mathcal{R}*})$ satisfies \eqref{eq:strong-dcbf-def}.
So, the feasible set of inputs is not reduced at any $x \in \mathcal{X}$.

\subsection{Optimization Formulation}
\label{subsec:optimization}

We adopt the philosophy of the NMPC-DCBF method to enforce safety constraints between polytopes, as shown in Sec. \ref{subsec:obs-avoidance-dual-variables}, and construct a multi-step optimization formulation.
For simplicity of notation, we drop the explicit dependence of $h_i(x_{t+k|t})$ on $x_{t|t}$ and $[u^T_{t|t},...,u^T_{t+k-1|t}]^T$.
Since $h_i(x_{t+k|t})$ is also not explicitly known at time $t$, to impose DCBF constraints along the horizon, we extend the idea in Sec. \ref{subsec:obs-avoidance-dual-variables} by enforcing a constraint stronger than $h_i(x_{t+k+1|t}) \geq \gamma_{k}h_i(x_{t+k|t})$ which does not rely on computations of $h_i(x_{t+k|t})$ and $h_i(x_{t+k+1|t})$ via \eqref{eq:min-dist-primal-qp}.

The primal optimization problem \eqref{eq:min-dist-primal-qp} provides an upper bound to $h_i(x)$, which can be used in the stronger DCBF constraints.
Let $(y^{\mathcal{O}_i},y^{\mathcal{R}})$ be any feasible solution of \eqref{eq:min-dist-primal-qp}.
Since \eqref{eq:min-dist-primal-qp} is a minimization problem and its solution is well-defined for all $x \in \mathcal{X}$,
\begin{equation}
\label{eq:prim-feas-ineq}
\bar{h}_i(x,y^{\mathcal{O}_i},y^{\mathcal{R}}) := \lVert y^{\mathcal{O}_i}-y^{\mathcal{R}} \rVert_2^2 \geq h_i(x).
\end{equation}
Then at time $t$, we can enforce
\begin{equation}
\label{eq:strong-ho-dcbf-def}
-\lambda^{\mathcal{O}_i} b^{\mathcal{O}_i} - \lambda^{\mathcal{R}} b^{\mathcal{R}}(x_{t+k+1|t}) \geq \gamma_k\lVert y^{\mathcal{O}_i}-y^{\mathcal{R}} \rVert_2^2
\end{equation}
which, using \eqref{eq:dual-feas-ineq} and \eqref{eq:prim-feas-ineq}, implies
\begin{equation}
\label{eq:strong-ho-dcbf-proof}
\begin{split}
    h_i(x_{t+k+1|t}) & \geq {-}\lambda^{\mathcal{O}_i} b^{\mathcal{O}_i} {-} \lambda^{\mathcal{R}} b^{\mathcal{R}}(x_{t+k+1|t}) \\
    & \geq \gamma_k\lVert y^{\mathcal{O}_i}-y^{\mathcal{R}} \rVert_2^2 \geq \gamma_kh_i(x_{t+k|t}),
\end{split}
\end{equation}
as required. Additionally, we can introduce the relaxation variables without affecting the analysis in this section.

\optional{
\begin{remark}
The primal problem \eqref{eq:min-dist-primal-qp} and the dual problem \eqref{eq:min-dist-dual-prob} provide upper and lower bounds respectively for the distance function $h$, which is implicit in nature.
These bounds can be calculated implicitly since they are equal to the cost functions subject to the corresponding constraints of each optimization problem.
Thus, any implicit constraint on $h$ can be converted into a weaker explicit set of constraints using these bounds.
\end{remark}
}

Then at time $t$, we first calculate the optimal solutions $(y^{\mathcal{O}_i*}_t,y^{\mathcal{R}*}_t)$ to the minimum distance QP \eqref{eq:min-dist-primal-qp} using $x=x_{t}$, hence the NMPC-DCBF formulation for polytopes is shown as follows:

\noindent\rule{\columnwidth}{0.4pt}
\textbf{NMPC-DCBF for Polytopes:}
\begin{subequations}
\label{eq:mod-mpc-dcbf}
\begin{align}
    & \min_{U, \Omega}p(x_{t+N|t}){+}\sum_{k=0}^{N-1}q(x_{t+k|t},u_{t+k|t}){+}\psi(\omega_k) \label{subeq:mod-mpc-dcbf-cost}\\
     \text{s.t.} \ & x_{t|t} = x_t, \quad y^{\mathcal{O}_i}_0 = y^{\mathcal{O}_i*}_t, \quad y^{\mathcal{R}}_0 = y^{\mathcal{R}*}_t \label{subeq:mod-mpc-dcbf-initial-condition} \\
    & x_{t+k+1|t} = f(x_{t+k|t}, u_{t+k|t}), \, k{=}0,...,N{-}1\label{subeq:mod-mpc-dcbf-dynamics-constraint} \\
    & u_{t+k|t} \in \mathcal{U}, \ x_{t+k|t} \in \mathcal{X}, \qquad k{=}0,...,N{-}1 \label{subeq:mod-mpc-dcbf-input-constraint} \\
    &{-}\lambda^{\mathcal{O}_i}_{k+1} b^{\mathcal{O}_i}{-}\lambda^{\mathcal{R}}_{k+1} b^{\mathcal{R}}(x_{t+k+1|t}) \geq \omega_k\gamma_k\lVert y^{\mathcal{O}_i}_k{-}y^{\mathcal{R}}_k \rVert_2^2, \notag \\
    & \qquad \text{for }k{=}0,...,N_{\text{CBF}}{-}1 \label{subeq:mod-mpc-dcbf-cbf-constraint} \\
    & A^{\mathcal{R}}(x_{t+k+1|t})y^{\mathcal{R}}_k \leq b^{\mathcal{R}}(x_{t+k+1|t}), \ A^{\mathcal{O}_i}y^{\mathcal{O}_i}_k \leq b^{\mathcal{O}_i}, \quad \;\; \notag \\
    & \qquad \text{for } k{=}1,...,N_{\text{CBF}}{-}1 \label{subeq:mod-mpc-dcbf-geo-cons} \\
    & \lambda^{\mathcal{O}_i}_{k} A^{\mathcal{O}_i}{+}\lambda^{\mathcal{R}}_{k} A^{\mathcal{R}}(x_{t+k|t}){=}0, \ \lVert \lambda^{\mathcal{O}_i}_{k} A^{\mathcal{O}_i} \rVert_2 \leq 1, \notag \\
    & \qquad \text{for } k{=}1,...,N_{\text{CBF}} \label{subeq:mod-mpc-dcbf-plane-cons} \\
    & \lambda^{\mathcal{O}_i}_{k+1} \geq 0, \quad \lambda^{\mathcal{R}}_{k+1} \geq 0, \quad \omega_k \geq 0 , \notag \\
    & \qquad \text{for } k{=}0,...,N_{\text{CBF}}{-}1 \label{subeq:mod-mpc-dcbf-non-neg-cons}
\end{align}
\end{subequations}
\noindent\rule{\columnwidth}{0.4pt}
\noindent
The subscripts of $\lambda^{\mathcal{O}_i}, \lambda^{\mathcal{R}}, y^{\mathcal{O}_i}, y^{\mathcal{R}}$ denote the time, \eqref{subeq:mod-mpc-dcbf-cbf-constraint} is the DCBF constraint between polytopes, \eqref{subeq:mod-mpc-dcbf-geo-cons} is the primal feasibility condition, and \eqref{subeq:mod-mpc-dcbf-plane-cons}-\eqref{subeq:mod-mpc-dcbf-non-neg-cons} are the dual feasibility conditions.
The initial conditions, system dynamics constraints, and input and state constraints are represented by \eqref{subeq:mod-mpc-dcbf-initial-condition}, \eqref{subeq:mod-mpc-dcbf-dynamics-constraint}, and \eqref{subeq:mod-mpc-dcbf-input-constraint} respectively.
This NMPC-DCBF formulation \eqref{eq:mod-mpc-dcbf} corresponds to enforcing safety constraints only between the pair $(\mathcal{O}_i, \mathcal{R})$ of polytopes.
To enforce DCBF constraints between every pair of robot and obstacle, we introduce corresponding dual and primal variables and enforce the constraints \eqref{subeq:mod-mpc-dcbf-cbf-constraint}-\eqref{subeq:mod-mpc-dcbf-non-neg-cons} for each pair.

\optional{
\begin{remark}
The mathematical intuition behind \eqref{eq:mod-mpc-dcbf} is different from the previous work in~\cite{thirugnanam2021duality} that focuses on duality in the continuous domain. Moreover, the system dynamics is not required to be control affine in our proposed \eqref{eq:mod-mpc-dcbf}. Furthermore, differentiability of dual variables is required in~\cite{thirugnanam2021duality} and not needed in this work.
\end{remark}
}

\begin{figure*}[!htp]
    \centering
    \begin{subfigure}[t]{0.24\linewidth}
        \centering
        \includegraphics[width = 0.99\linewidth]{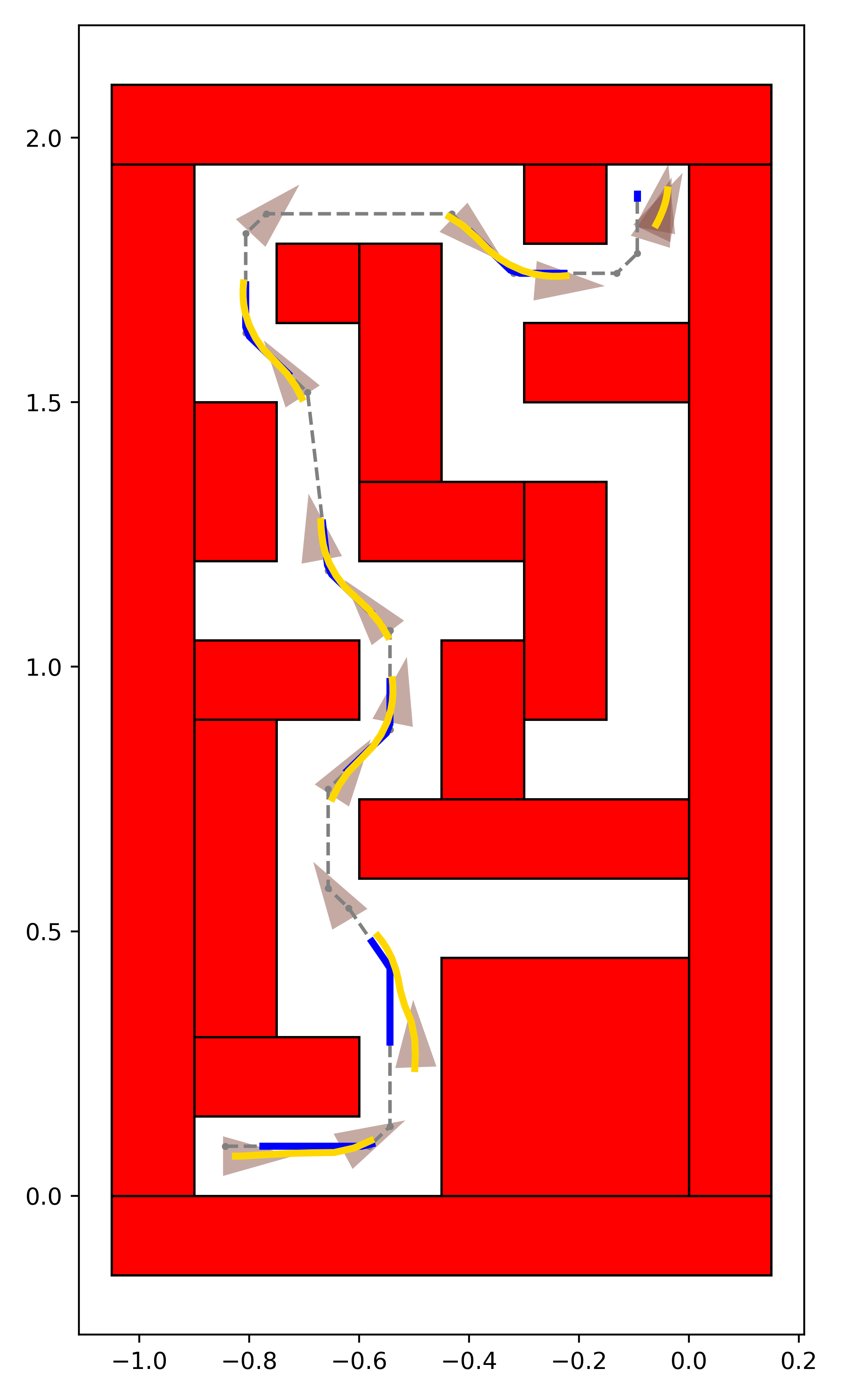}
        \caption{Triangle-shaped robot}
        \label{subfig:snapshots-triangle1}
    \end{subfigure}
    \begin{subfigure}[t]{0.24\linewidth}
        \centering
        \includegraphics[width = 0.99\linewidth]{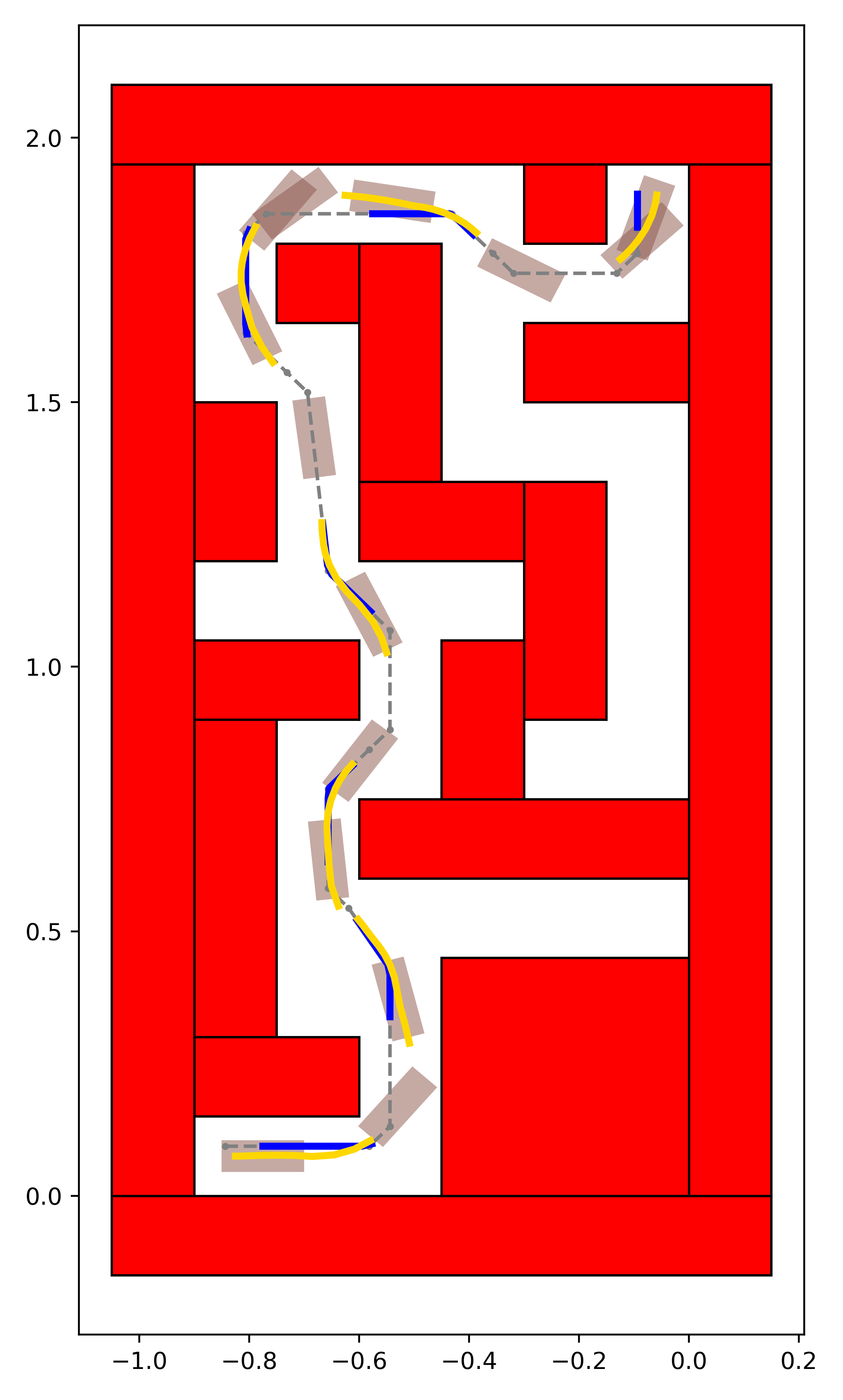}
        \caption{Rectangle-shaped robot}
        \label{subfig:snapshots-rectangle1}
    \end{subfigure}
    \begin{subfigure}[t]{0.24\linewidth}
        \centering
        \includegraphics[width = 0.99\linewidth]{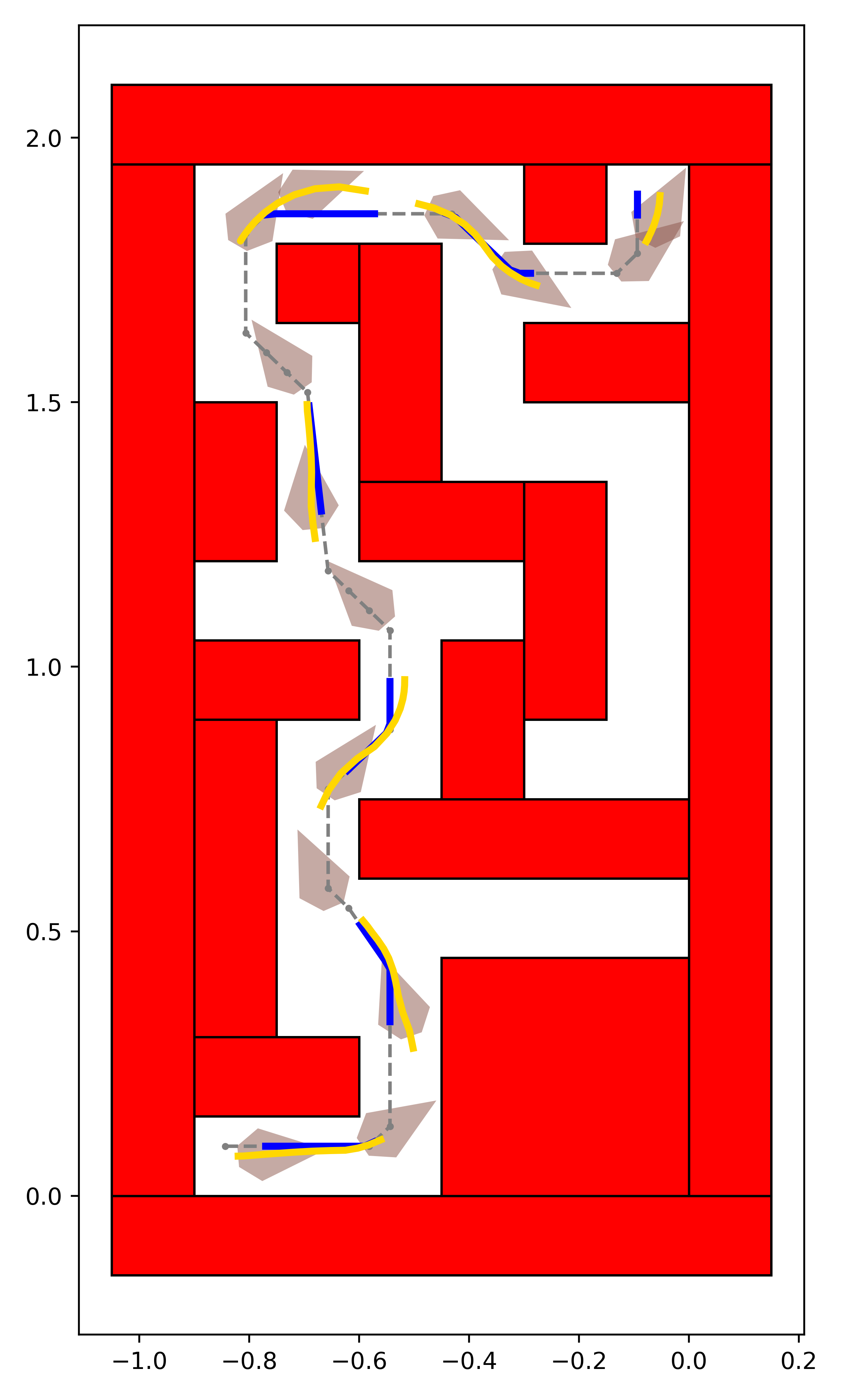}
        \caption{Pentagon-shaped robot}
        \label{subfig:snapshots-pentagon1}
    \end{subfigure}
    \begin{subfigure}[t]{0.24\linewidth}
        \centering
        \includegraphics[width = 0.99\linewidth]{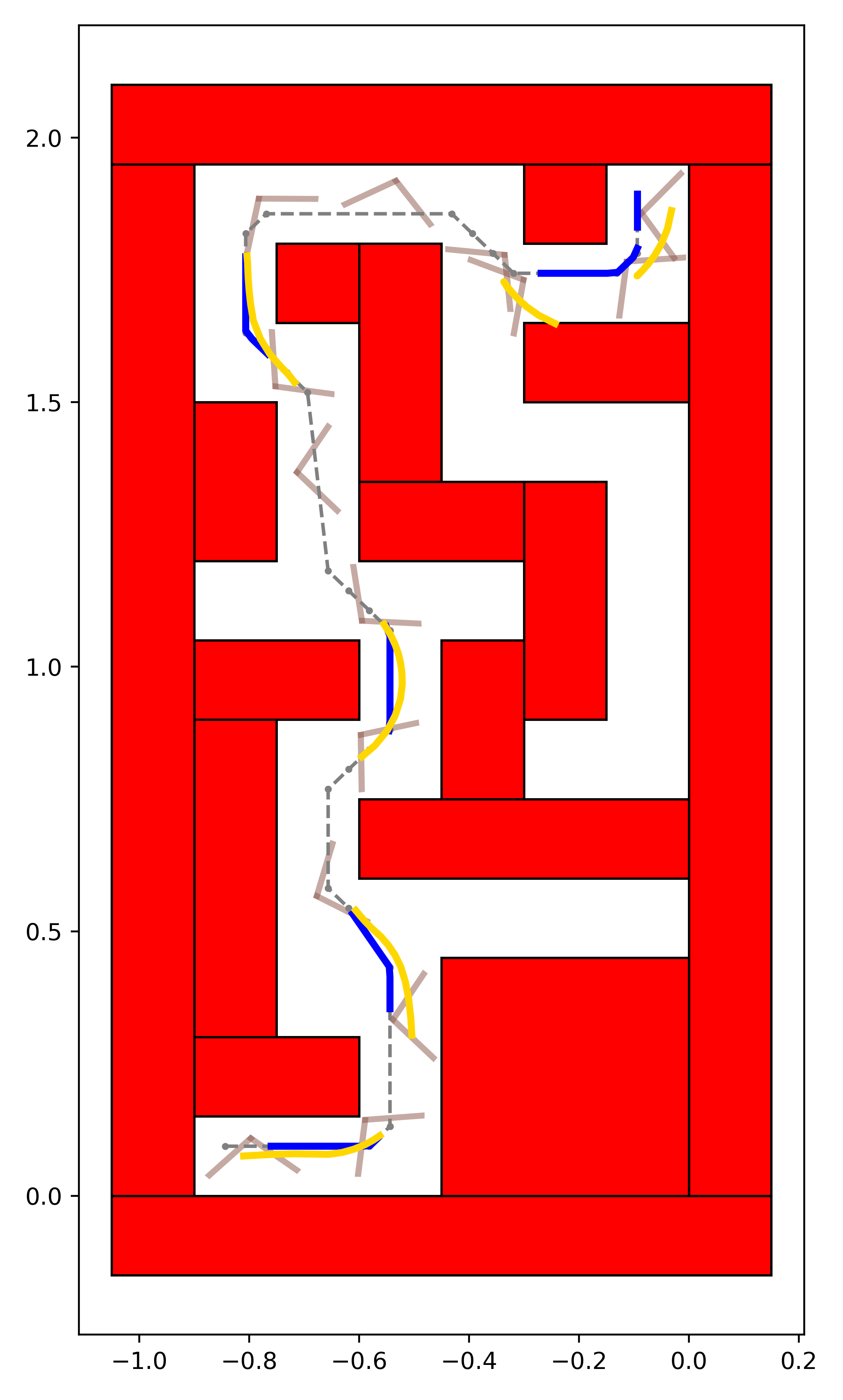}
        \caption{L-shaped robot}
        \label{subfig:snapshots-lshape1}
    \end{subfigure} \\
    \centering
    \begin{subfigure}[t]{0.24\linewidth}
        \centering
        \includegraphics[width = 0.99\linewidth]{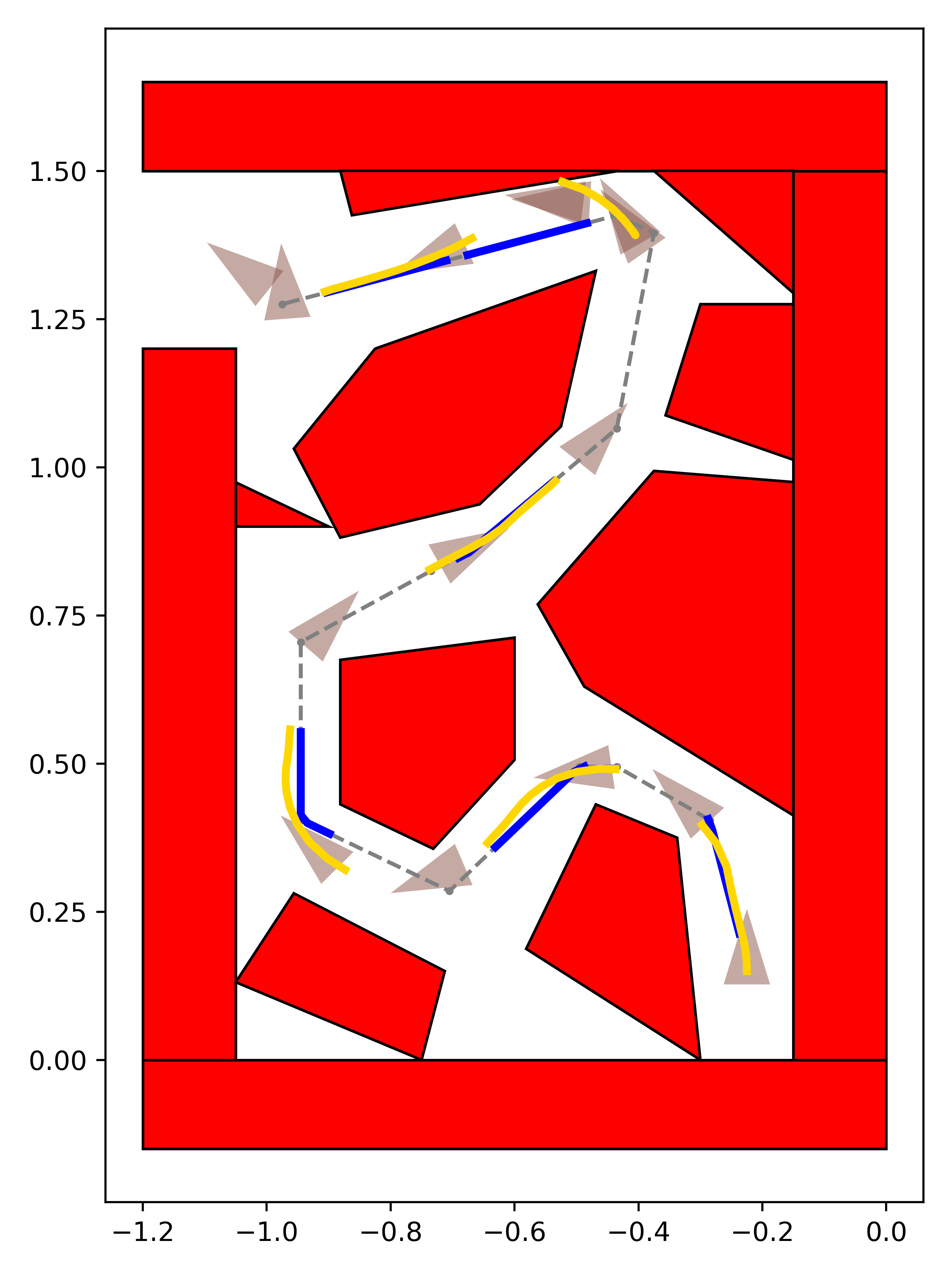}
        \caption{Triangle-shaped robot}
        \label{subfig:snapshots-triangle2}
    \end{subfigure}
    \begin{subfigure}[t]{0.24\linewidth}
        \centering
        \includegraphics[width = 0.99\linewidth]{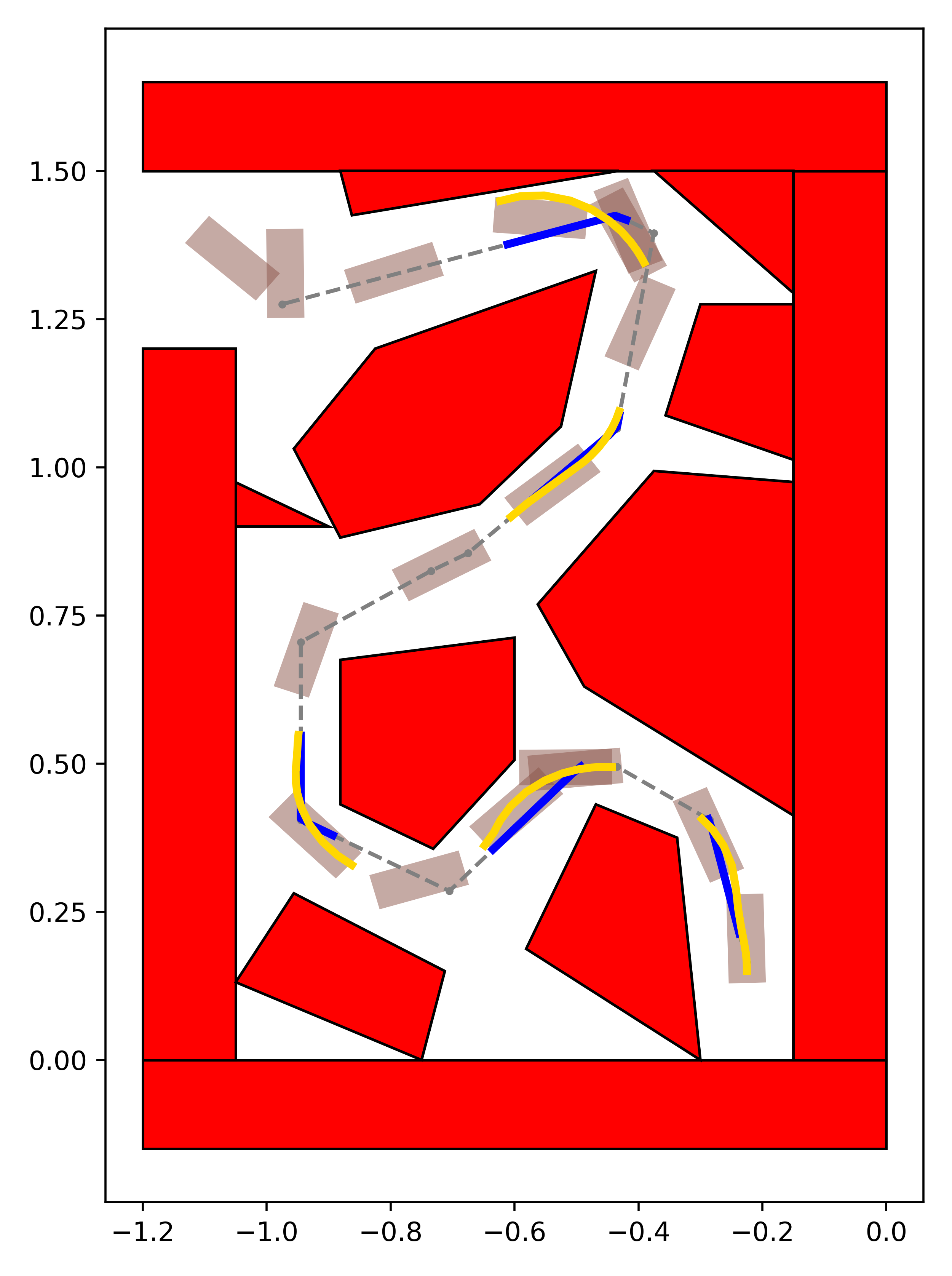}
        \caption{Rectangle-shaped robot}
        \label{subfig:snapshots-rectangle2}
    \end{subfigure}
    \begin{subfigure}[t]{0.24\linewidth}
        \centering
        \includegraphics[width = 0.99\linewidth]{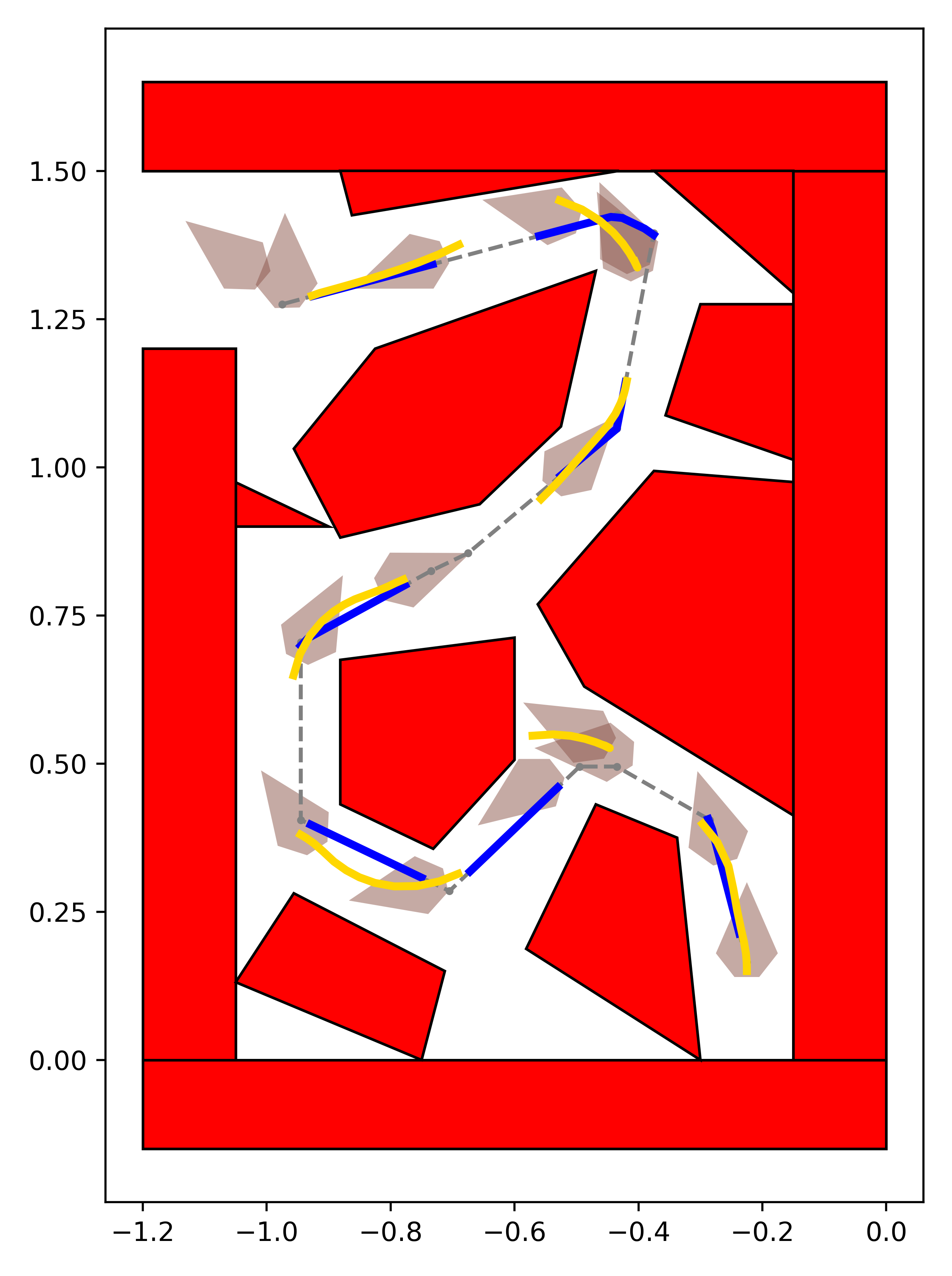}
        \caption{Pentagon-shaped robot}
        \label{subfig:snapshots-pentagon2}
    \end{subfigure}
    \begin{subfigure}[t]{0.24\linewidth}
        \centering
        \includegraphics[width = 0.99\linewidth]{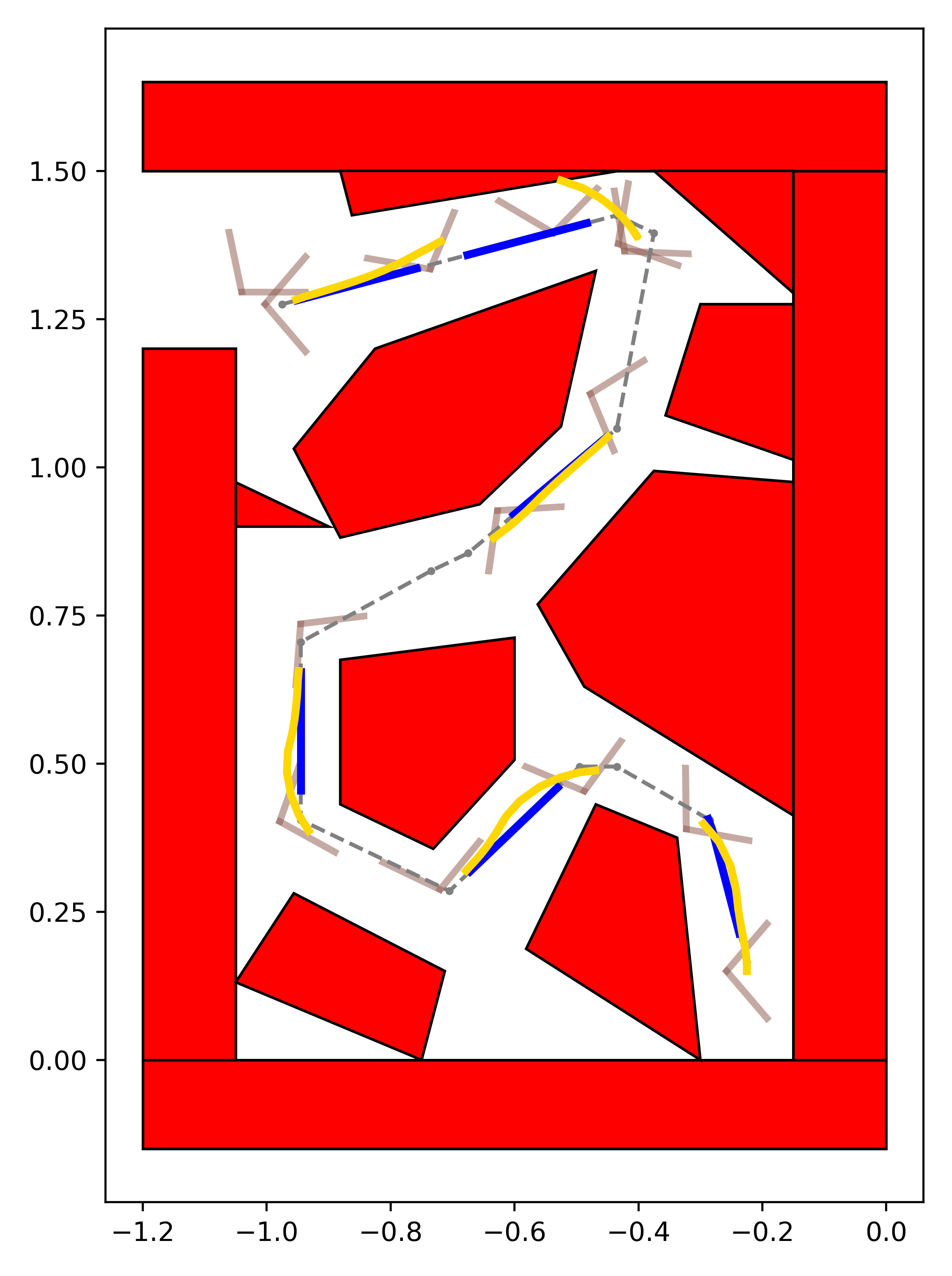}
        \caption{L-shaped robot}
        \label{subfig:snapshots-lshape2}
    \end{subfigure}
    \caption{
    Snapshots from simulation of tight maneuvers of obstacle avoidance with a controlled robot with different shapes in two maze environments.
    The triangle has side lengths $0.133 m, 0.133m, 0.1m$; the rectangle has length $0.15 m$ and width $0.06 m$; the symmetric pentagon has length $0.16 m$ and width $0.1 m$; and the L-shape has an arm length $0.114 m$, arm width $0.03 m$, with included angle of $101 \deg$.
    The corridors in both the maze environments have widths $0.15 m$, and every robot geometry has at least one dimension larger than the corridor width.
    The grey dotted lines represent the global reference path, the blue lines represent the local reference trajectory for the optimization, and the yellow lines represent the optimized trajectory at various time instances.}
    \label{fig:snapshots}
\end{figure*}

\subsection{Complexity and Performance}
\label{subsec:complexity-performance}
\subsubsection{Exponential DCBF Constraint}
To reduce complexity of the NMPC-DCBF shown in \eqref{eq:mod-mpc-dcbf}, we can modify the DCBF constraint $h_i(x_{t+k|t}) \geq \omega_k \gamma_{k} h_i(x_{t+k-1|t})$ by rolling out time $k$ and removing the dependence on $x_{t+k-1|t}$ from the LHS of the DCBF constraint, thus enforcing the following exponential DCBF constraints:
\begin{equation}
\label{eq:alt-ho-dcbf-def}
h_i(x_{t+k|t}) \geq \omega_k (\Pi_{j=0}^{k} \gamma_{j}) h_i(x_{t|t}).
\end{equation}
The exponential decay rate $\Pi_{j=0}^{k} \gamma_{j}$ arises due to rolling out the decay rate at time $j$, $\gamma_j$.
The DCBF constraint \eqref{eq:alt-ho-dcbf-def} only contains $h_i(x_{t|t})$ on the RHS for all $k$, which can be explicitly computed at each time step using $x_t = x_{t|t}$.
Note that the RHS of the DCBF constraint \eqref{eq:strong-ho-dcbf-proof} at time $k\geq 1$ cannot be computed explicitly, since $x_{t+k|t}$ implicitly depends on the control inputs.
This leads to modifying \eqref{subeq:mod-mpc-dcbf-cbf-constraint} with,
\begin{equation}
    -\lambda^{\mathcal{O}_i}_{k} b^{\mathcal{O}_i}-\lambda^{\mathcal{R}}_{k} b^{\mathcal{R}}(x_{t+k|t}) \geq \omega_k (\Pi_{j=0}^{k} \gamma_{j}) h_i(x_{t|t})
\end{equation}
Such a formulation speeds up the computational time, as only $h(x_{t|t})$ needs to be calculated at each time $t$.
This change affects neither the feasibility nor the safety of the system.

\subsubsection{Horizon Length Selection}
\label{subsubsec:horizon-length-dcbf}
Compared with distance constraints, DCBF constraints allow effective obstacle avoidance behavior with smaller horizon length.
Our NMPC-DCBF formulation does not require the obstacle avoidance horizon length $N_{\text{CBF}}$ be equal to $N$, which additionally reduces the complexity~\cite{zeng2021enhancing}.
Additionally, maneuvers such as deceleration for obstacle avoidance or reversing motion for deadlock avoidance are motion primitives that only require a small horizon in the MPC formulation.
To sum up, DCBF constraints with dual variables enable fast optimization for obstacle avoidance between polytopes.



\section{Numerical Results}
\label{sec:results}
In this section, we consider an autonomous navigation problem.
We model the controlled robot with different shapes, including triangle, rectangle, pentagon and L-shape. 
The proposed optimization-based planning algorithm allows us to successfully generate dynamically-feasible collision-free trajectories even in tight maze environments, shown in Fig.~\ref{fig:snapshots}.
Animation of the navigation problems can be found in the video attachment.

\subsubsection{Environment}
The tight maze environment is described by a combination of multiple convex obstacles, including shapes like triangle, rectangle pentagon, etc.
The controlled robot is also modelled with different shapes, including triangle, rectangle, pentagon and L-shape, whose orientation is determined by the yaw angle.
The L-shape is non-convex, but can be represented by two convex polytopes.
The dimensions for each robot shape are mentioned in Fig.~\ref{fig:snapshots}.

\subsubsection{System Dynamics}
The controlled robot is described by the kinematic bicycle model, which is typical for testing trajectory planning algorithms in tight environments. The continuous-time dynamics of the robot is given as follows,
\begin{equation}
\label{eq:kinematic-bicycle-model}
\dot{c}_x = v \cos(\phi), \ \dot{c}_x = v \sin(\phi), \ \dot{v} = a, \ \dot{\phi} = \dfrac{v \tan{\delta}}{l},
\end{equation}
where system states are $x = (c_x, c_y, v, \phi)$ with $(c_x, c_y)$ as the center of rear axes, $\phi$ as yaw angle and $v$ as the velocity, and $u = (a, \delta)$ are inputs with steering angle $\delta$ and acceleration $a$. The wheel base of the robot is $l=0.1m$.
The steering angle and acceleration are limited between $\pm 0.5rad$ and $\pm 1 m/s^2$.

\subsubsection{Global Planning}
The global path from the starting position to the goal is generated using the A$^*$ algorithm.
The 2-D space is sub-divided into grids and obstacle collision checks are performed at each grid point during the algorithm.
A safety margin, which is smaller than at least one dimension of the robot, is used for the collision checks.
Finally, the generated optimal path is reduced to fewer waypoints using line-of-sight reductions, similar to the $\theta^*$ algorithm~\cite{nash2007theta}.
The generated global path is not dynamically feasible, and is only safe at the node points.

\subsubsection{Local Trajectory Generation}
The local trajectory planning is formulated using the NMPC-DCBF formulation \eqref{eq:mod-mpc-dcbf} to track the local reference trajectory while avoiding obstacles.
The local reference trajectory $\bar{X} = [\bar{x}^T_{t|t}, \bar{x}^T_{t+1|t}, ..., \bar{x}^T_{t+N|t}]^T$ is generated from a start point with a constant speed $v_0$ and the same orientation as the global planner.
The start point is found by local projection from the current robot's position to the global path.
For tracking the local reference trajectory, the cost function \eqref{eq:mod-mpc-dcbf} in the optimizer is constructed with terminal cost, stage cost, and relaxing cost function, respectively as
\begin{subequations}
\label{eq:mpc-dcbf-cost-functions}
\begin{align}
    p(x_{t+N|t}) = & || x_{t+N|t} - \bar{x}_{t+N|t}||^2_{Q_T} \label{subeq:mpc-dcbf-cost-functions-terminal-cost}, \\
    q(x_{t+k|t}, u_{t+k|t}) = & || x_{t+k|t} - \bar{x}_{t+k|t} ||^2_{Q} + ||u_{t+k|t}||^2_{R} \label{subeq:mpc-dcbf-cost-functions-stage-cost} \\ \nonumber
    & + ||u_{t+k|t} - u_{t+k-1|t}||^2_{dR} , \\
    \psi(\omega_k) = & p_{\omega} (\omega_k - 1)^2 \label{subeq:mpc-dcbf-cost-functions-terminal-cost-relaxation},
\end{align}
\end{subequations}
where $u_{t-1|t} = u^*_{t-1|t-1}$ represents the last optimized control input.
The dynamics constraints \eqref{subeq:mod-mpc-dcbf-dynamics-constraint} is applied with the discrete-time forward Euler formulation from the continuous-time dynamics \eqref{eq:kinematic-bicycle-model}.
The input constraints \eqref{subeq:mod-mpc-dcbf-input-constraint} is imposed by the steering angle and acceleration limits mentioned above.
A prediction horizon of $N = 11$ is used, with the DCBF horizon as $N_{\text{CBF}} = 6$, and the decay rate as $\omega_k = 0.8$.
The obstacle avoidance constraints \eqref{subeq:mod-mpc-dcbf-cbf-constraint}-\eqref{subeq:mod-mpc-dcbf-non-neg-cons} can be applied directly to each pair of the convex-shaped (triangle, rectangle, pentagon) robot and each convex obstacle. When the robot is non-convex shaped (L-shape), these constraints are applied to the convex parts of the robot and each convex obstacle.

To reduce the complexity of the optimization formulation, we only consider the obstacles which are within a specified radius from the robot at any given time.
The radius is calculated using the reference tracking velocity, prediction horizon and the maximum deceleration of the robots.

\subsubsection{Warm Start}
The NMPC-DCBF formulation is a non-convex optimization, and hence computationally challenging to solve in general.
Although the DCBF constraints help to reduce the complexity, as discussed in Sec. \ref{subsubsec:horizon-length-dcbf}, it still requires a good initial guess to lead to faster numerical convergence.
The initial guess trajectory and control inputs are generated using a braking controller.
An acceleration input equal to the maximum deceleration is provided and the steering angle is set to zero.
Once the robot comes to a halt, the acceleration input is also set to zero.
The braking control inputs, along with the trajectory generated from it are provided as an initial guess to the optimization at each time step.
Since at each time step $h_i(x_{t|t})$ is solved using \eqref{eq:min-dist-dual-prob}, the dual optimal solution from this computation is provided as an initial guess for the dual variables for the entire horizon.

\subsubsection{Simulation Results}
To evaluate the performance of \eqref{eq:mod-mpc-dcbf}, we study the navigation problem with two different maze environments with four choices of robot shapes.
The optimization problems are implemented in Python with CasADi~\cite{andersson2019casadi} as modelling language, solved with IPOPT~\cite{biegler2009large} on Ubuntu 18.04 with Intel Xeon E-2176M CPU with a 2.7GHz clock.
From the snapshots we can observe tight-fitting obstacle avoidance motion of the robot, and also reversing motion to avoid deadlock.
These examples highlight the safety and planning features of our implementation.
We also analyze the computational time of trajectory generation using \eqref{eq:mod-mpc-dcbf}, which is shown in TABLE \ref{tab:computational-time}.
This illustrates that optimization \eqref{eq:mod-mpc-dcbf} can be solved sufficiently fast to be deployed on different-shaped robots for trajectory generation in different maze environments.
The details of hyperparameter selections can be found in the open-source repository.

The specific choice of the parameters in the optimization formulation \eqref{eq:mod-mpc-dcbf} can influence safety and deadlock behaviors in the robot.
Qualitatively, for safety, the reference tracking velocity should be such that the robot can come to a halt within the prediction horizon with the input as the maximum deceleration.
There is also a trade\edit{-}off between deadlock avoidance and safety: Higher value of the terminal cost weight $Q_T$ improves deadlock avoidance, but can also increase velocity of the robot\edit{,} leading to unsafe motion.

\optional{
To tune the hyperparameters for a given system, consider the set of feasible trajectories of \eqref{eq:mod-mpc-dcbf} for some reference trajectory $\bar{X}$.
First, to capture the affect of all control inputs on the DCBF $h$, the safety horizon $N_{\text{CBF}}$ must be larger than the relative degree of the system with respect to $h$.
If there exists any safe trajectory at the current state that brings the robot to rest, the resulting control law guarantees that the closed-loop trajectory is also safe.
Thus, it is desirable to choose the safety horizon $N_{\text{CBF}}$ high enough so that the robot can come to rest within $N_{\text{CBF}}$ steps.
However, choosing a large safety horizon may result in an increased computation time and safety could be achieved in practice with proper tuning of the decay rate $\gamma$.
Choosing $\gamma$ closer to $1$ prioritizes safety over reference tracking while choosing $\gamma$ closer to $0$ ensures faithful tracking at the cost of safety.
Note that the relaxation variable $\omega_k$ is essential for feasibility of \eqref{eq:mod-mpc-dcbf} especially for $\gamma$ closer to $1$, see~\cite{zeng2021enhancing}.
Selecting a large terminal cost parameter $p_\omega$ in \eqref{subeq:mpc-dcbf-cost-functions-terminal-cost-relaxation} is also key to ensure that the effective decay rate does not deviate, thus prioritizing safety of the system.
Finally, the stage cost parameter $Q$ in \eqref{subeq:mpc-dcbf-cost-functions-stage-cost} and the terminal cost parameter $Q_T$ in \eqref{subeq:mpc-dcbf-cost-functions-terminal-cost} present a trade-off between reference tracking and exploration.
Larger values of $Q$ result in the closed-loop trajectory closely following the reference trajectory.
Notice that since the reference trajectory may not be dynamically feasible, this may still result in deadlocks.
Selecting larger values of $Q_T$ mostly penalizes deviation from the terminal state and thus promotes exploration during timesteps $k=1,...,N_{\text{CBF}-1}$.
This can result in better deadlock avoidance maneuvers as depicted in Fig.~\ref{fig:snapshots}.
}

\begin{table}[!htp]
\caption{Solver time statistics of NMPC-DCBF with polytopes \eqref{eq:mod-mpc-dcbf}.}
\centering
\begin{tabular}{c|l|cccc}
\hline
Env & Robot Shape & median & std & min & max \\ \hline
\multirow{4}{*}{Maze 1} & triangle (Fig.~\ref{subfig:snapshots-triangle1}) & 51ms & 19ms & 14ms  & 149ms  \\
 & rectangle (Fig.~\ref{subfig:snapshots-rectangle1}) & 49ms  & 25ms  & 15ms & 185ms \\
 & pentagon (Fig.~\ref{subfig:snapshots-pentagon1}) & 71ms & 15ms & 15ms & 272ms \\
 & L-shape (Fig.~\ref{subfig:snapshots-lshape1}) & 85ms & 13ms & 13ms & 215ms \\ \hline
\multirow{4}{*}{Maze 2} & triangle (Fig.~\ref{subfig:snapshots-triangle2}) & 33ms & 24ms & 13ms  & 121ms \\
 & rectangle (Fig.~\ref{subfig:snapshots-rectangle2}) & 29ms & 25ms & 13ms & 122ms \\
 & pentagon (Fig.~\ref{subfig:snapshots-pentagon2}) & 30ms & 29ms & 13ms & 150ms \\
 & L-shape (Fig.~\ref{subfig:snapshots-lshape2}) & 29ms & 49ms & 13ms & 233ms \\
 \hline
\end{tabular}
\label{tab:computational-time}
\end{table}
\section{Conclusion}
\label{sec:conclusion}
We proposed a nonlinear optimization formulation using discrete-time control barrier function based constraints for polytopes.
The proposed formulation has been shown to be applied as a fast optimization for control and planning for general nonlinear dynamical systems.
We validated our approach on navigation problems with various robot shapes in maze environments with polytopic obstacles.

\balance
\bibliographystyle{IEEEtran}
\bibliography{references}{}

\end{document}